\documentclass[12pt]{article}
\usepackage[a4paper,margin=1in]{geometry}
\usepackage{graphicx}
\usepackage{amsmath}
\usepackage{xcolor}
\definecolor{blue}{rgb}{0,0,0}
\usepackage{natbib}
\usepackage{hyperref}
\hypersetup{colorlinks=true, allcolors=blue}
\usepackage{tikz}
\usetikzlibrary{arrows.meta,positioning,fit,calc}
\usepackage{array}
\usepackage{makecell}
\newcolumntype{C}[1]{>{\centering\arraybackslash}p{#1}}

\title{Agentic AI and Retrieval-Augmented Models in Straight-Through Underwriting}
\author{Robert Richardson, Josh Meyers, Brian Hartman, and David Sandberg \\ Brigham Young University}
\date{}

\begin{document}

\maketitle

\begin{abstract}
Artificial intelligence (AI) is beginning to reshape actuarial practice, particularly in domains that require reasoning over unstructured documents, heterogeneous data sources, and regulated decision workflows. Actuaries now face a design space that ranges from traditional rule-based automation to large language models (LLMs), retrieval-augmented generation (RAG), and multi-agent ``agentic'' systems that plan, retrieve, call tools, and reflect. This paper examines how these emerging architectures can support actuarial priorities such as transparency, auditability, and human-in-the-loop governance, with a focus on straight-through decision processes. To make these ideas concrete, we develop and analyze an agentic AI framework for straight-through underwriting of small commercial Business Owner Policies (BOPs). We construct a synthetic but realistic experimental environment and compare three underwriting pipelines: (i) a single-LLM baseline, (ii) a naive RAG system, and (iii) a multi-agent ``Agentic RAG'' pipeline that combines targeted retrieval, third-party data checks, and explicit multi-step rule evaluation.  The agentic system performs best overall, with the largest gains in multi-step and missing-information scenarios, where structured retrieval and reflection help the model avoid unsupported straight-through decisions.
\end{abstract}

\textit{Keywords:} Retrieval-Augmented Generation (RAG); Human-AI Interactions; 
 Synthetic Data Generation; Insurance Automation;

\section{Introduction}

Artificial intelligence (AI) has undergone rapid evolution in recent years, moving from rule-based automation toward models with increased adaptability and autonomy. Traditional automation systems execute predefined workflows, often lacking flexibility when confronted with unstructured or unforeseen data. In contrast, large language models (LLMs) such as those powered by transformer architectures have enabled richer capabilities: they can process unstructured natural language inputs, generate fluent text, and perform a variety of downstream tasks. 

\textcolor{blue}{The distinction between rule-based automation and LLM-based systems should not be interpreted too sharply. In many applications, explicit rules, statistical models, and LLMs often play complementary roles. A rule engine is well suited to apply known decision logic once the relevant facts are available. An LLM, by contrast, may be useful when those facts are embedded in messy business descriptions, underwriting notes, or other unstructured text. In some cases, LLMs may also help surface recurring patterns or exceptions that can later be reviewed by domain experts and translated into explicit underwriting rules. The practical question is therefore not whether rules or LLMs are better, but how each can be used in the parts of the workflow where it is most reliable.}

One of the pivotal developments in this space is the technique known as retrieval-augmented generation (RAG). In the foundational work by \citet{lewis2020rag}, RAG models are defined as combining a parametric memory with a non-parametric retrieval mechanism. The authors show that such a hybrid architecture improves performance on knowledge-intensive tasks by enabling the system to access external factual content rather than relying solely on internal model parameters.  Subsequent surveys and reviews reinforce that RAG architectures help mitigate phenomena like ``hallucination'' (where the model asserts facts not grounded in the external world) and improve factuality and traceability \citep{gupta2024survey}.

While RAG enhances knowledge access, it remains a relatively narrow paradigm: retrieval plus generation. The next frontier is what many authors term ``agentic AI,'' which are systems that behave more like autonomous agents, capable of setting sub-goals, invoking external tools, coordinating workflows, and even collaborating across multiple agents. Throughout this discussion, ``agentic'' does not imply sentience or true autonomy, but rather structured orchestration of statistical models. Recent taxonomy distinguishes multi-agent collaboration, task decomposition, memory, and coordinated autonomy as key features of agentic architectures \citep{botti2025agentic}. In this view, agentic systems go beyond simply responding to prompts. They exhibit what the literature calls ``reasoning'' about how to achieve higher-level goals---again, a statistical planning heuristic rather than genuine human reasoning. Importantly, throughout this paper, we use terms such as ``reasoning'' or ``thinking'' in the sense adopted by the AI research community. These systems do not think or reason in the human cognitive sense; rather, they estimate outputs conditioned on learned statistical structure. Their effectiveness remains dependent on careful human oversight and domain expertise.

In the domain of actuarial science where decisions often involve uncertain data, regulatory constraints, interpretability, and risk modelling, there is an opportunity to deploy these agentic frameworks. Traditional actuarial tasks such as underwriting, reserving, claims processing, and risk monitoring are increasingly data-rich and complex. By integrating retrieval, LLM-based inferential heuristics, structured statistical models, and tool orchestration into an agentic pipeline, one can imagine underwriting systems that not only process applications but generate structured, auditable steps in support of actuarial decision-making.

\textcolor{blue}{The motivation for using an LLM in this setting is not that it replaces traditional underwriting rules. Many underwriting decisions still depend on clear thresholds or exclusions that can be applied directly once the relevant facts are known. The challenge is that those facts are often not presented cleanly. They may appear in a business narrative, be implied by several fields, or be missing from the application but available in another source. The LLM component is used to help map this unstructured information to the relevant guidebook concepts, summarize missing or ambiguous facts, and produce a draft rationale. The RAG and agentic components then place structure around that language capability by grounding the output in retrieved guidebook text, requiring structured decisions, and routing uncertain cases to human review.}

This paper presents an exploration of such an agentic AI framework applied to straight-through underwriting in a business owner policy (BOP) insurance setting. After defining the conceptual distinctions between automation, LLMs, retrieval-augmented generation, and agentic AI, we review agentic use cases in actuarial science, discuss synthetic/generated data pipelines, describe our specific framework, and present empirical results. We conclude with implications, limitations, and directions for future work.

\section{Agentic AI Concepts}

Artificial intelligence has evolved through distinct stages of capability, from rule-based automation to generative and ultimately agentic systems. Each stage represents an expansion of what constitutes ``intelligence'' in practical systems, from executing predefined logic to performing parameter-driven reasoning heuristics with goals, tools, and memory. These advances remain computational and statistical in nature; human judgment remains essential for interpreting outputs responsibly.

\subsection{Large Language Models (LLMs)}

Early automation systems followed fixed decision rules. In business processes such as underwriting or claims handling, these systems relied on deterministic logic designed to replicate human procedures, such as if-then trees, linear scorecards, and expert systems \citep{russell2016ai}. They excelled at efficiency and repeatability but failed when encountering data or scenarios outside their programmed domain. 

Machine learning extended automation by allowing systems to learn parameters from data rather than explicit instructions. Supervised models such as logistic regression and gradient-boosted trees offered statistical inference but not genuine reasoning or contextual awareness.

Large Language Models represent a significant leap from classical automation. Based on transformer architectures \citep{vaswani2017attention}, they learn latent structure in language, enabling flexible generation, summarization, and what the AI literature refers to as ``reasoning.'' These are not acts of cognition but probabilistic pattern extrapolations. LLMs remain parametric, meaning all information is encoded in their weights, making them prone to hallucination \citep{ji2023surveyhallucination}. 

In practical terms, naive LLM deployments behave as sophisticated text generators, capable of producing coherent narratives but lacking direct access to verified facts, structured databases, or organizational knowledge. Their value is realized only when paired with rigorous human review and well-designed workflows.

\subsection{Retrieval-Augmented Generation}

Retrieval-Augmented Generation (RAG) addresses key limitations of purely parametric language models by coupling an LLM with a non-parametric memory \citep{lewis2020rag}. In a RAG pipeline, the model:
\begin{enumerate}
\item Encodes a user query into a dense vector representation.
\item Searches a vector database of domain documents for semantically similar content.
\item Injects the retrieved passages into the LLM prompt.
\item Generates an output grounded in this external context.
\end{enumerate}

This architecture enhances factual accuracy and interpretability by ensuring each response can be traced to specific supporting documents.  
For regulated domains such as finance or healthcare, this grounding is critical for compliance, auditability, and explainability \citep{gupta2024survey}.  

\textbf{Limitations.}  
Despite its strengths, RAG systems are not without drawbacks.  
Performance depends heavily on the quality of document retrieval---errors in search or embedding similarity can cause the model to operate over irrelevant or outdated passages \citep{asai2023selfrag, izacard2022few}.  
Moreover, LLMs may still hallucinate or misattribute facts even when provided with relevant context \citep{ji2023survey}.  
RAG pipelines are also reactive: they answer a single query without persistent memory, planning, or reflection, limiting their ability to refine or validate their inferential process over multiple steps.  
In high-stakes applications such as actuarial or financial modeling, these shortcomings must be managed through human judgment.

\textbf{Recent Enhancements.}  
Several approaches aim to mitigate these issues.  
Self-RAG architectures introduce self-reflective verification, prompting models to critique and re-rank retrieved passages before generation \citep{asai2023selfrag}.  
Dynamic retrieval and query rewriting methods update search queries iteratively to reduce context drift \citep{trivedi2023rethinking}.  
Memory-augmented RAG systems further maintain persistent state, allowing multiple rounds of retrieval and reasoning \citep{shuster2022blenderbot3}.  
These improvements begin to blur the boundary between retrieval-based and agentic reasoning systems.

\subsection{Agentic AI}

Agentic AI extends beyond retrieval to orchestration.  
According to recent frameworks from Microsoft's \emph{AutoGen} project and open research in multi-agent reasoning \citep{li2023autogen,botti2025agentic}, an agentic system is characterized by autonomy, tool use, memory, and reflective reasoning.  
Instead of a single prompt--response cycle, it maintains a stateful interaction loop:
\begin{enumerate}
\item The agent interprets the user goal (statistically, by estimating intent from input text).
\item It plans sub-tasks and invokes external tools (retrievers, calculators, APIs, models).
\item It reflects on intermediate results, adapting subsequent actions.
\item It terminates when the goal is achieved or uncertainty thresholds are met.
\end{enumerate}

\textcolor{blue}{In this paper, we use the term ``agentic'' in a narrow and practical sense. The system is not autonomous in the human sense, and it does not reason in the way an underwriter reasons. Instead, it organizes a set of defined workflow steps: retrieving guidebook evidence, routing the application into accept, reject, or refer pathways, checking whether decisive facts are missing, using simulated third-party data when available, applying multi-step underwriting rules, and reviewing the final decision. This narrower definition is important because the value of the system comes from making the decision process more structured and inspectable, not from giving the model independent authority.}

This framework embeds LLM-based inference within a control architecture, often implemented with orchestration libraries such as LangChain or LangGraph, where the model not only generates text but also decides \emph{when} to retrieve, compute, or delegate.  
Because this behavior is algorithmic rather than cognitive, effective use depends critically on smart human design, validation, and supervision.

\textbf{Limitations.}  
While promising, multi-agent systems introduce new points of failure.  
Communication errors, inconsistent context passing, and cascading prompt noise can lead to instability or non-deterministic behavior across agents \citep{wang2024survey}.  
Each additional agent adds complexity to alignment and safety verification, making the system more difficult to audit or reproduce \citep{park2023generative}.  
Furthermore, multi-agent orchestration often incurs higher computational cost and latency, and reflective loops may amplify biases if not properly constrained.

\textbf{Emerging Mitigations.}  
Recent work addresses these vulnerabilities through improved coordination and verification mechanisms.  
For instance, delegation-aware architectures constrain inter-agent communication to predefined schemas \citep{liu2024agentbench}.  
Critic--executor frameworks employ specialized ``critic'' agents that evaluate and revise outputs from task agents, improving reliability and stability \citep{madaan2024selfrefine}.  
Other research focuses on memory management and role consistency, ensuring agents retain coherent goals across iterative interactions \citep{chen2024mindagent}.  
Together, these refinements aim to balance autonomy with control---an essential requirement in professional domains such as actuarial science, where traceability and reproducibility are mandatory.

\section{AI in Actuarial Science}

The actuarial profession is increasingly incorporating advanced artificial intelligence (AI) capabilities. These approaches are reshaping how actuaries approach reserving, pricing, and risk monitoring by enabling parameter-driven ``reasoning'' over unstructured data, dynamic workflows, and semi-autonomous decision pipelines. Their advantages, however, materialize only when skilled actuaries design, monitor, and interpret these systems.

Several studies and professional reports highlight how actuaries and insurers are already applying AI methods. A foundational literature review by the Society of Actuaries (SOA) emphasized that AI and machine learning will ``transform actuarial work through the creation of new technologies, products, and services'' \citep{soa2019literaturereview}. More recently, the SOA's \emph{Primer on Generative AI for Actuaries} recommends that generative systems be deployed as augmentation tools for actuarial judgment rather than full automation \citep{soageneratorai2024}. Industry analysis from Oliver Wyman further reports that AI in insurance is already enhancing accuracy, fraud detection, and customer service, while introducing new data-governance challenges \citep{balona2023actuarygpt}. 

Recent work titled \emph{Advanced Applications of Generative AI in Actuarial Science} provides concrete demonstrations of RAG and multi-agent systems performing document comparison, vision-assisted claim classification, and automated reporting \citep{hatzesberger2025generativeai}. 
The Casualty Actuarial Society has called for research on leveraging large language models (LLMs) to process unstructured claims data for actuarial use \citep{richman2024aivision}.  
An AI-based automated pricing and underwriting model has been demonstrated for the general insurance sector, integrating machine learning with rule-based decision processes \citep{mahohoho2024automatedactuarial}.  
Case studies highlight LLMs deriving features from text, RAG automating document retrieval, and agentic multi-step systems orchestrating data ingestion and statistical inference \citep{hatzesberger2025generativeai}.  
The application of AI in actuarial science is new and underdeveloped, but active research is underway to determine how these tools can best support rather than replace professional actuarial expertise.

\subsection{Use Cases of Agentic AI in Actuarial Practice}
\label{sec:use-cases}

Agentic AI should be viewed not as a substitute for actuarial expertise but as an amplifier of it.
Its strength lies in coordinating retrieval, tool use, and reflection, which are activities that are procedural yet time-consuming, while preserving professional judgment and accountability.
\textcolor{blue}{The following four scenarios are intended as illustrative use cases rather than predictions of immediate production capability. In practice, the feasibility of each workflow depends on data quality, system integration, governance constraints, and the degree to which legacy platforms can expose reliable inputs to the agentic system.}

\subsubsection*{Scenario 1: Pricing Cycle Management and Regulatory Filing Support}
\textbf{Context.} Actuaries performing quarterly rate reviews must reconcile exposure and loss data, run rate indications, prepare exhibits, and document the rationale for regulatory filings.
These cycles blend data engineering, modeling, and narrative justification, guided by the \emph{CAS Statement of Principles on Ratemaking} \citep{cas2019statement}.

\textbf{Agentic augmentation.}
\begin{itemize}
\item \textcolor{blue}{\textbf{Workflow-assistance tools} could help extract data from internal warehouses, populate portions of indication templates, and identify reconciliation issues in earned exposures, premiums, and losses, provided that source systems, data dictionaries, and schema mappings have been validated.}
\item \textcolor{blue}{\textbf{Agentic review modules} could provide preliminary checks of pipeline outputs, examining assumptions for consistency and reasonableness across exhibits (e.g., trend selections, development patterns, inflation factors), while leaving actuarial selections and final judgment to the practitioner.}
\item \textcolor{blue}{\textbf{Reflection components} could draft explanatory exhibits or generate diagnostic code (e.g., SQL queries) to help investigate period-over-period changes, such as shifts in written premium trends. These outputs would require review, especially when source databases are complex, poorly documented, or maintained across multiple legacy systems.}
\item \textcolor{blue}{\textbf{Retrieval systems} could compare assumptions and methodologies against prior filings, competitor submissions, and state bulletins to help identify possible deviations or emerging regulatory considerations \citep{balona2023actuarygpt}.}
\item \textcolor{blue}{\textbf{Draft summaries} with data provenance, assumption documentation, and cross-references to supporting materials could be assembled for actuarial review and sign-off.}
\end{itemize}

\textbf{Benefits.} \textcolor{blue}{When the underlying data connections are reliable, the combination of automated data preparation and agentic review may improve consistency, reduce some manual coordination, and support more complete documentation.}
Actuaries retain full authority over rate indications and filing decisions, \textcolor{blue}{and the system should be treated as a documentation and review aid rather than an independent filing engine.}

\textbf{Risks.} Overreliance on automated diagnostics may obscure data quality issues, regulatory nuance, or business context that actuaries are better positioned to interpret.
\textcolor{blue}{These risks are especially important in insurance environments with legacy databases, inconsistent coding practices, complex product hierarchies, or multiple downstream reporting systems.}
Agentic modules may also surface false positives or misleading comparisons, requiring careful human validation to avoid unjustified adjustments or misaligned filings.

\subsubsection*{Scenario 2: Reserve Diagnostics and Portfolio Monitoring}
\textbf{Context.} Reserving actuaries analyze loss development across lines and methods, explain movements, and communicate results to finance, claims, and regulatory stakeholders.
Classical methods such as the chain-ladder, Bornhuetter--Ferguson, and stochastic reserving approaches \citep{mack1993accuracy,england2002stochastic,wuthrich2008stochastic} remain foundational, yet diagnostic synthesis and documentation remain labor-intensive.

\textbf{Agentic augmentation.}
\begin{itemize}
\item \textcolor{blue}{\textbf{Automated pipeline checks} could help schedule data refreshes, execute pre-specified reserving techniques (e.g., chain-ladder, Bornhuetter--Ferguson, GLM- or Bayesian-based stochastic methods), and compare outputs across valuation periods, subject to existing data controls and actuarial review.}
\item \textcolor{blue}{\textbf{Variance analysis agents} could draft preliminary movement explanations by analyzing changes in loss triangles and identifying possible drivers such as large-loss emergence, development pattern shifts, or operational changes (e.g., ``Increase in paid severity for AY 2021 $\rightarrow$ +3.2 points IBNR'').}
\item \textcolor{blue}{\textbf{RAG-based claim linking} could help search individual claim notes and transaction histories to identify claims associated with unusual cell-level movements, while recognizing that claim narratives may be incomplete, inconsistent, or affected by operational coding practices.}
\item \textcolor{blue}{\textbf{Assumption monitoring modules} could compare current selections (e.g., tail factors, a priori loss ratios, case reserve adequacy assumptions) against prior memos and methodological documentation to flag material deviations beyond established tolerances \citep{hatzesberger2025generativeai}.}
\item \textcolor{blue}{\textbf{Governance logs} could store model outputs, diagnostic notes, assumption lineage, and human overrides to support reproducibility, audit trails, and internal model governance.}
\end{itemize}

\textbf{Benefits.} \textcolor{blue}{The actuary could receive a structured diagnostic aid that highlights areas requiring expert judgment rather than a fully automated reserving conclusion.}
Interpretability, communication consistency, and governance discipline \textcolor{blue}{may improve} while methodological choices remain fully under actuarial control.

\textbf{Risks.} Automatically generated explanations may overstate confidence or misattribute drivers of development, especially in sparse or volatile triangles.
Agentic systems may also focus on statistically detectable changes while missing operational insights apparent to experienced reserving actuaries, reinforcing the need for human interpretation. \textcolor{blue}{For this reason, generated movement explanations should be treated as hypotheses to investigate, not as final reserve diagnostics.}

\subsubsection*{Scenario 3: Claims Triage and Straight-Through Processing Support}
\textbf{Context.} P\&C insurers pursue straight-through processing (STP) for simple claims while maintaining rigorous oversight of complex or high-risk ones.
Actuaries contribute by designing triage and severity models that balance efficiency with fairness \citep{richman2024aivision}.

\textbf{Agentic augmentation.}
\begin{itemize}
\item \textcolor{blue}{The agent could retrieve policy terms, prior claims, and adjuster notes to contextualize new submissions, subject to data availability and access controls.}
\item \textcolor{blue}{Reasoning modules could apply triage rules and invoke fraud or severity scoring models, while preserving the distinction between model indicators and claim-handling decisions.}
\item \textcolor{blue}{Reflection steps could compare outputs with historical experience and escalate uncertain cases with documented rationale.}
\item \textcolor{blue}{Transparent logs could show which rules, models, and data sources informed each recommendation.}
\end{itemize}

\textbf{Benefits.} \textcolor{blue}{The framework could support faster handling of low-complexity claims while improving documentation and auditability.}
Actuaries retain control of model calibration and validation but gain enhanced visibility into operational performance.

\textbf{Risks.} Automated triage may overweight efficiency at the expense of fairness if not carefully monitored; actuaries are typically better at identifying equity considerations, claim-specific nuance, and unintended model bias.
Escalation logic may also be brittle, requiring oversight to avoid inappropriate straight-through approvals or rejections. \textcolor{blue}{Because claims decisions can have direct policyholder impact, any straight-through workflow should include monitoring for error patterns, disparate impact, and inappropriate automation of borderline cases.}

\subsubsection*{Scenario 4: Catastrophe Exposure and Accumulation Management}
\textbf{Context.} Catastrophe (CAT) modeling integrates hazard, exposure, vulnerability, and financial modules to estimate loss distributions \citep{grossi2005catastrophe}.
Actuaries oversee validation, communication of PML/TVaR, and governance per industry best practices \citep{guycarp2023cat}.

\textbf{Agentic augmentation.}
\begin{itemize}
\item \textcolor{blue}{A planning agent could help coordinate exposure updates and vendor model runs when the required data feeds and model interfaces are sufficiently standardized.}
\item \textcolor{blue}{Retrieval components could pull prior exposure reports, treaty terms, and hazard maps to support review of model inputs and assumptions.}
\item \textcolor{blue}{Reflection layers could flag unusual exposure growth or potential concentration issues for actuarial review, producing annotated dashboards where appropriate.}
\item \textcolor{blue}{When model results exceed pre-specified thresholds, the system could draft alerts with location-level summaries and comparable-event references.}
\end{itemize}

\textbf{Benefits.} \textcolor{blue}{Actuaries may spend less time assembling routine inputs and more time interpreting results, testing assumptions, and communicating uncertainty.}
Agentic orchestration \textcolor{blue}{could improve governance by preserving clearer records of inputs, model versions, assumptions, and review steps.}

\textbf{Risks.} Excessive reliance on automated workflows may shift focus toward process compliance rather than the substantive quality of scenario interpretation.
Actuaries must ensure that automated anomaly detection does not overshadow expert judgment regarding exposure growth, data limitations, and model uncertainty. \textcolor{blue}{This is especially important because CAT results are sensitive to exposure quality, vendor model assumptions, and the treatment of low-probability high-severity events.}

\subsubsection{Additional Opportunities for Augmentation}
Beyond these four cases, agentic orchestration \textcolor{blue}{may also support selected components of}:
\begin{itemize}
\item \textbf{Reinsurance analytics:} \textcolor{blue}{retrieving treaty wordings, summarizing key provisions, and supporting scenario comparisons.}
\item \textbf{Operational data quality:} \textcolor{blue}{helping coordinate reconciliations between policy, billing, and claims systems, while flagging exceptions for human review.}
\item \textbf{Regulatory and financial reporting:} \textcolor{blue}{drafting disclosure narratives that can be checked against prior filings and current assumptions.}
\item \textbf{Portfolio steering:} \textcolor{blue}{monitoring loss-ratio drift and identifying emerging exposure clusters for further analysis.}
\item \textbf{Agent / Broker Monitoring:} \textcolor{blue}{reviewing changes in mix of business from agents or brokers and flagging patterns that may warrant underwriting review.}
\end{itemize}

\textcolor{blue}{Across these use cases, the most realistic near-term role for agentic AI is to reduce repetitive coordination work, improve documentation, and surface issues for professional review. Its value depends on careful implementation, reliable data infrastructure, and actuarial oversight consistent with the profession's standards of accuracy, fairness, and accountability.}

\subsection{Ethical Considerations}

\label{sec:ethics}

Ethical reflection is central to the responsible adoption of artificial intelligence in actuarial work.
The actuarial profession has long emphasized professionalism, accountability, and the public interest, while AI ethics frameworks stress fairness, transparency, and human oversight.
As actuaries deploy large language models, retrieval systems, and agentic frameworks, both traditions must converge to ensure responsible practice.

\textcolor{blue}{This governance framing is consistent with current practitioner concerns in actuarial and insurance settings. Guidance from actuarial and insurance organizations generally emphasizes that AI tools should support, rather than replace, professional judgment. For this framework, that means the system should preserve the information needed for review: the data used, the guidebook passages retrieved, the model output, the reason for any escalation, and any human override. The goal is not only to improve accuracy, but also to make the workflow easier to inspect, challenge, and govern. Because regulatory expectations for AI continue to evolve, this paper should be read as a proof of concept for a governance-aware workflow rather than as a deployment-ready compliance framework.}

\subsubsection{Alignment with Actuarial Professional Standards}
Actuarial ethics, as codified in the \emph{CAS Code of Professional Conduct} and the \emph{SOA Code of Professional Conduct}, require actuaries to act with integrity, competence, and due care, using appropriate models and methods.
Precept~3 requires that any model used in professional work be fit for purpose and adequately communicated.
AI and agentic systems introduce new responsibilities: validating algorithmic components, disclosing limitations, and ensuring that automated reasoning supports and not supplants human judgment.
These align with model-risk-management expectations under the NAIC's \emph{Model Risk Management Handbook} (2023) and emerging AI governance guidance.

From a broader AI ethics perspective, five recurring principles appear across major frameworks such as the OECD AI Principles and the NAIC \emph{Principles on Artificial Intelligence} (2020):
\emph{(1)} fairness and non-discrimination,
\emph{(2)} accountability and human oversight,
\emph{(3)} transparency and explainability,
\emph{(4)} privacy and security, and
\emph{(5)} robustness and reliability.
For actuarial applications, these translate into clear documentation of data sources and assumptions, interpretability of decision logic, and audit trails that regulators and policyholders can trust.
Scholarly reviews note that responsible AI systems require proportional human involvement, outcome monitoring, and ongoing impact assessment \citep{jobin2019global, floridi2018ai4people}.
Such principles are consistent with actuarial professionalism's focus on verifiable methods and communication of uncertainty.

\subsubsection{Practical Integration in Actuarial Workflows}
In practice, ethical AI governance for actuarial systems should include:
\begin{itemize}
\item \textbf{Human-in-the-loop design:} ensuring actuaries retain authority for acceptance, override, and sign-off decisions.
\item \textbf{Model transparency:} maintaining documentation that explains algorithms, data provenance, and limitations.
\item \textbf{Bias and fairness testing:} routinely checking for disparate impact across demographic or geographic segments.
\item \textbf{Auditability and traceability:} storing reasoning traces, retrieval evidence, and model outputs for regulatory review.
\item \textbf{Continuous validation:} treating AI components as models subject to annual review under model-risk frameworks.
\item \textcolor{blue}{\textbf{Implementation controls:} documenting data interfaces, system dependencies, known data-quality limitations, and circumstances in which automated outputs should not be relied upon without additional review.}
\end{itemize}

Ultimately, the ethical integration of AI within actuarial science should rest on a shared foundation:
AI ethics contributes principles of responsible autonomy and accountability, while actuarial professionalism provides the structures of peer review, disclosure, and the public-interest mandate.
An agentic system that enhances transparency, reduces bias, and maintains human oversight can thus be viewed as a natural extension of actuarial ethics rather than a disruption of it.

\section{Synthetic Data Generation and the Business Owner Policy (BOP) Use Case}

\subsection{Use Case Motivation and Context}

The motivating problem for this work is the automation of straight-through underwriting (STU) for small commercial Business Owner Policies (BOPs). In conventional workflows, underwriting requires actuaries, analysts, and underwriters to interpret lengthy policy manuals, evaluate diverse data sources, and decide whether an application should be accepted, rejected, or referred for human review. The process is information-intensive, and small differences in how facts are interpreted can change the final decision.

The envisioned agentic pipeline uses:
\begin{itemize}
  \item A large language model (LLM) to read and summarize application materials.
  \item A retrieval component to ground decisions in the synthetic underwriting guidebook.
  \item A structured orchestration layer that checks completeness, retrieves third-party information when needed, decomposes multi-step rules, and escalates cases when the decisive information is unavailable.
\end{itemize}

\textcolor{blue}{Testing such a pipeline requires data that reflect business diversity, missing-information patterns, edge cases, and underwriting rules that may depend on information distributed across several parts of an application. Real commercial underwriting data cannot generally be shared for regulatory, privacy, and competitive reasons, and it rarely includes explicit ground-truth rationales. At the same time, a synthetic BOP dataset necessarily falls short of a production underwriting or pricing environment. Real BOP workflows may involve hundreds of variables, company-specific rating plans, legacy data systems, detailed coverage forms, and complex interactions among classification, exposure, payroll, revenue, premises characteristics, loss history, and operational controls. Consequently, our synthetic dataset should be interpreted as a controlled proof-of-concept environment rather than a full representation of production BOP underwriting.}

\textcolor{blue}{Within that limitation, the dataset was designed to preserve several important logical and structural properties of real underwriting decisions. The applications include complete cases, explicit rule violations, missing-information cases, and multi-step reasoning cases in which the correct decision depends on connecting information from different parts of the application, the underwriting guide, and simulated third-party data. For example, an application may report total square footage in one field, describe that only a portion of the premises is used for a particular operation in the narrative, and require comparison with a guidebook rule that applies only if that operation exceeds a specified square-footage threshold. Other cases require interpreting operating hours, employee counts, revenue limits, or conditional exclusions from messy business descriptions. These examples do not make the dataset fully realistic, but they create a transparent benchmark for evaluating whether agentic retrieval and review steps improve decision accuracy and explanation quality relative to simpler LLM and RAG baselines.}

This dataset allows us to address several research challenges:
\begin{enumerate}
  \item \textbf{Incomplete information:} Many applications omit details essential for risk evaluation.
  \item \textbf{Multiple decision pathways:} A case can fail because of a direct appetite restriction, a multi-step rule violation, or unresolvable ambiguity.
  \item \textbf{Need for consistent reasoning:} Evaluation requires not only the correct outcome but also a defensible explanation.
  \item \textbf{Integration of heterogeneous data:} Underwriting decisions depend on structured fields, unstructured narratives, guidebook clauses, and third-party information.
\end{enumerate}

By including simulated third-party data, rule-based guidebook clauses, and multi-step reasoning cases, the synthetic dataset enables end-to-end testing of the entire agentic underwriting pipeline, from data ingestion and retrieval to decision justification and human escalation.
\color{black}

\subsection{Policy Guidebook Generation}

We first generated a 143-page synthetic policy guidebook using a large language model. The guidebook encodes general underwriting standards and business-specific rules for 127 business types. Each business type has its own appetite language, including eligibility criteria, exclusion criteria, and thresholds involving property characteristics, operational details, revenue mix, or business activities. In total there are 16 global requirements and 6 business-specific requirements for each business type. We recognize that the guidebook does not reproduce any insurer's internal underwriting manual, but it provides a controlled and realistic basis for generating applications that should be accepted, rejected, or referred. \textcolor{blue}{The full guidebook can be found at \url{https://github.com/drbob-richardson/agentic-bop-underwriting/blob/main/data/BOP_Guidebook.pdf}.}

Example clause:
\begin{quote}
``For restaurants serving alcohol, coverage is excluded if more than 50\% of revenue is derived from alcohol sales, unless a liquor-liability rider is attached.''
\end{quote}

This guidebook forms the primary retrieval corpus for the RAG and agentic components. In the multi-step scenarios, the relevant clause is often not sufficient by itself. The model must also extract or infer the numerical quantity or operational condition needed to apply the rule.

\subsection{Application and Third-Party Data Generation}

For each of the 127 business types, we created five application scenarios, producing 635 applications in total. Each scenario represents a distinct truth condition:

\begin{enumerate}
  \item \textbf{Compliant case:} The application contains the necessary information and meets all relevant guidebook criteria (\textit{Accept}).
  \item \textbf{Single-issue violation:} The application contains one explicit breach of the guidebook rules, such as a prohibited operation or a threshold violation stated directly in the application (\textit{Reject}).
  \item \textcolor{blue}{\textbf{Multi-step reasoning:} The correct decision requires combining at least two facts before applying the guidebook rule. The decisive issue may involve a derived square footage, revenue share, activity mix, subcontractor count, or shared-premises condition (\textit{Accept} or \textit{Reject}, depending on the derived result).}
  \item \textbf{Missing information (recoverable):} Key information is omitted from the application but can be recovered from simulated third-party data. The model must retrieve the missing information and then apply the guidebook rule.
  \item \textbf{Missing information (irrecoverable):} The decisive information is absent from both the application and the third-party data, so the proper outcome is \textit{Refer to human review}.
\end{enumerate}

Each record combines structured elements such as business class, location, revenue, employee count, property size, and prior losses with unstructured text such as narrative operations descriptions, underwriting notes, and third-party business summaries. The third-party data emulate external information sources that an agentic system could query in practice, such as public business descriptions, marketing pages, property summaries, or registry-style records.

Every rejection or referral scenario is paired with a consistent ground-truth rationale to support evaluation of explanation quality. For instance:

\begin{quote}
\textbf{True reason:} Reject because the application reports 14,000 total square feet, the operations note states that 50\% is used for grocery sales, and the guidebook permits grocery operations only below 5,000 square feet. The derived grocery area is 7,000 square feet, exceeding the threshold.\\
\textbf{Model rationale example:} ``The business should be declined because its grocery operation occupies approximately 7,000 square feet, which exceeds the 5,000-square-foot eligibility limit in the underwriting guide.''
\end{quote}

Evaluation compares the model's textual explanation to the true rationale using cosine-similarity scoring, enabling quantitative assessment of reasoning alignment in addition to decision accuracy.

A human validation team manually reviewed the synthetic applications, guidebook excerpts, and expected outcomes. Reviewers verified that each record's data, decision label, and rationale were logically consistent and realistic. This step prevented circular validation, because the systems were not evaluated solely against unreviewed AI-generated labels. The final dataset contains 127 applications in each of the five scenario categories. After validation and relabeling, the final decision outcomes are:
\begin{itemize}
  \item 127 applications that should be accepted (20.0\%).
  \item 345 applications that should be rejected (54.3\%).
  \item 163 applications that should be referred to human review (25.7\%).
\end{itemize}
\textcolor{blue}{The final dataset can be found at \url{https://github.com/drbob-richardson/agentic-bop-underwriting/blob/main/BOP_Synthetic_Dataset.csv}.}

\subsection{Examples}

To illustrate the structure of the  synthetic dataset, consider a small specialty grocery and cafe risk. The guidebook includes the following simplified clauses:
\begin{quote}
``Specialty grocery operations are eligible only when the grocery portion occupies less than 5,000 square feet. Cafe operations are eligible if seating is limited and alcohol sales are not offered. Shared premises with excluded manufacturing operations require referral unless the occupancy separation is documented.''
\end{quote}
These clauses create several possible decision paths.

\paragraph{(1) Compliant Case.}
\textit{Application excerpt:} ``Maple Street Cafe operates a 2,800-square-foot coffee and sandwich shop with no alcohol sales and no shared manufacturing or warehouse operations.'' All material facts are present and within appetite. \textbf{Decision:} \textit{Accept.}

\paragraph{(2) Single-Issue Violation.}
\textit{Application excerpt:} ``Maple Street Cafe offers evening service and derives 60\% of revenue from wine and beer sales.'' The alcohol condition is directly stated and violates the guidebook rule. \textbf{Decision:} \textit{Reject.}

\textcolor{blue}{\paragraph{(3) Multi-Step Reasoning.}
\textit{Application excerpt:} ``Maple Street Market leases 12,000 square feet. The front half of the premises is used for specialty grocery sales, with the remaining space used for office and storage.'' The guidebook threshold applies to the grocery portion, not the total premises. The model must compute that 50\% of 12,000 square feet is 6,000 square feet, then compare that derived amount with the 5,000-square-foot grocery limit. \textbf{Decision:} \textit{Reject.}}

\paragraph{(4) Missing Information (Recoverable).}
\textit{Application excerpt:} ``Maple Street Market operates a neighborhood grocery and cafe.'' The application does not state square footage by operation. Third-party property data state that the business leases 8,000 square feet and a public description states that the grocery area occupies one-third of the premises. The derived grocery area is approximately 2,667 square feet, which is below the threshold. \textbf{Decision:} \textit{Accept} if all other guidebook conditions are satisfied.

\paragraph{(5) Missing Information (Irrecoverable).}
\textit{Application excerpt:} ``Maple Street Market operates a neighborhood grocery and cafe in a mixed-use building.'' Neither the application nor third-party data disclose the grocery square footage or whether the premises are separated from other occupancies. The decisive information is unavailable. \textbf{Decision:} \textit{Refer to human review.}

These examples show why the revised experiment is harder than simple rule lookup. In the multi-step and missing-information cases, the model must determine what fact is needed, locate or derive that fact, and then decide whether the guidebook rule can be applied safely.

\section{Agentic Pipeline and Experimental Results}

\subsection{Model Descriptions}

The main purpose of the AI system is to determine whether a BOP application should be accepted, rejected, or referred to human review. The decision is based on whether the application is within the underwriting appetite described in the synthetic guidebook. In the revised experiment, there is no logistic-regression or quantitative risk-score layer. Instead, the third scenario category tests whether the model can apply multi-step underwriting logic: extracting relevant facts, deriving any needed intermediate quantity, retrieving the correct rule, and applying that rule consistently.

To put the agentic workflow into context, we evaluate three systems, from simplest to most structured. Conceptually, the goal of the agentic pipeline is not to replace underwriting judgment but to triage work. The AI should handle the easiest high-confidence passes and fails, while routing nuanced or low-confidence applications to human underwriters. Finding this balance requires careful calibration of prompts, retrieval rules, and escalation thresholds against human decisions.

\subsubsection{Single-LLM Baseline}

\begin{figure}[h!]
\centering
\includegraphics[width=\linewidth]{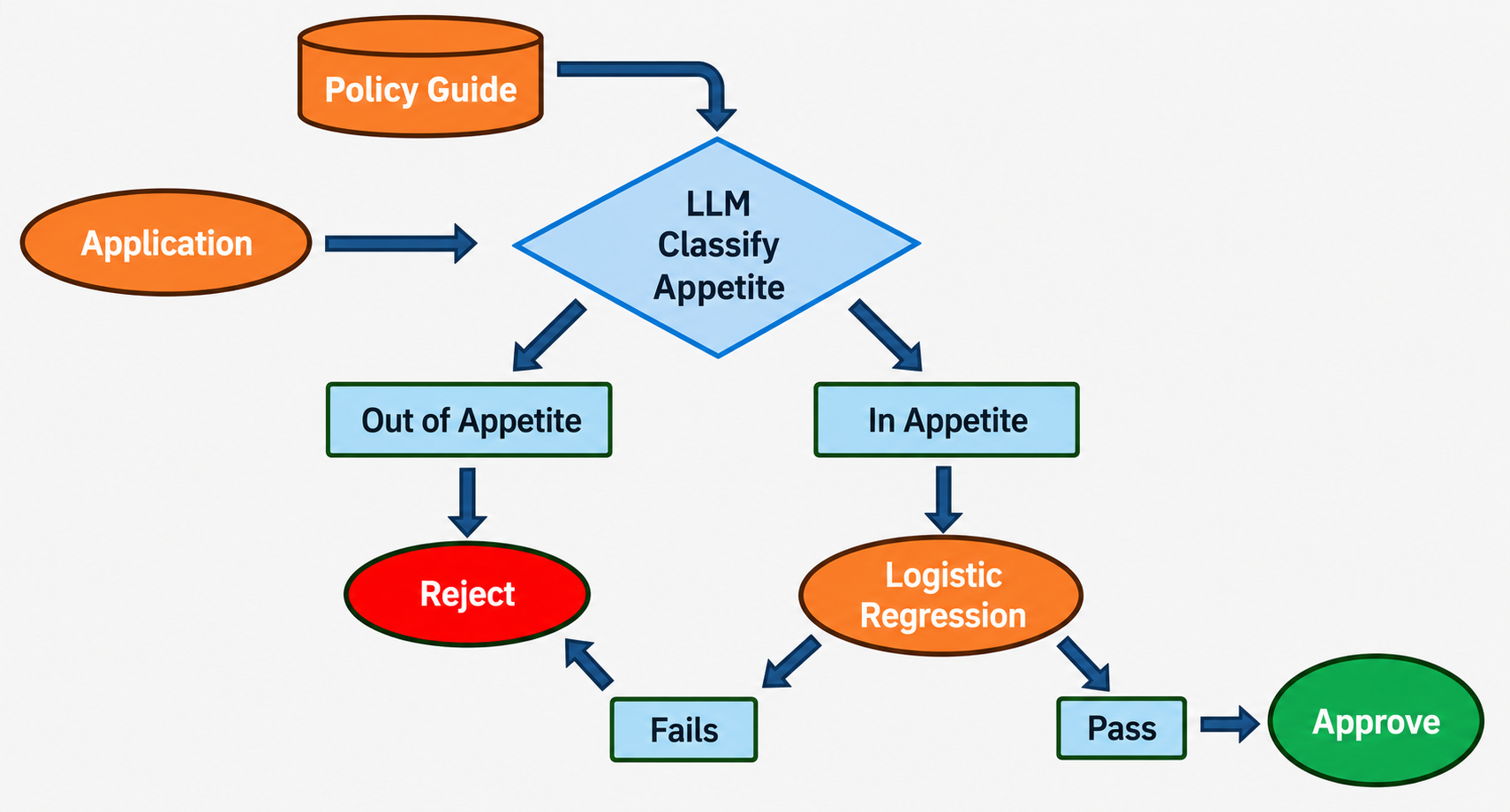}
\caption{\textbf{Single-LLM Baseline.} A single LLM determines the underwriting disposition directly from application text and static guidebook context. Decisions rely on a fixed prompt and do not use dynamic retrieval, explicit completeness checks, third-party lookup, or reflection.}
\label{fig:llm_workflow}
\end{figure}

Figure~\ref{fig:llm_workflow} shows the simplest underwriting workflow, in which a single large language model performs end-to-end decision-making based on the application data and static policy-guide context. We use the term ``Determine Appetite'' because it is the underwriting-appropriate term for the task, but again emphasize that the LLM is a parameter-based model and does not reason or think in the human sense.

In this configuration, the model assesses appetite directly from the provided text. Although simple and fast, this baseline lacks a structured mechanism for resolving ambiguous or missing-context cases. It may also miss multi-step violations when the decisive fact must be derived rather than read directly.

\subsubsection{Retrieval-Augmented Workflow (RAG)}

\begin{figure}[h!]
\centering
\includegraphics[width=\linewidth]{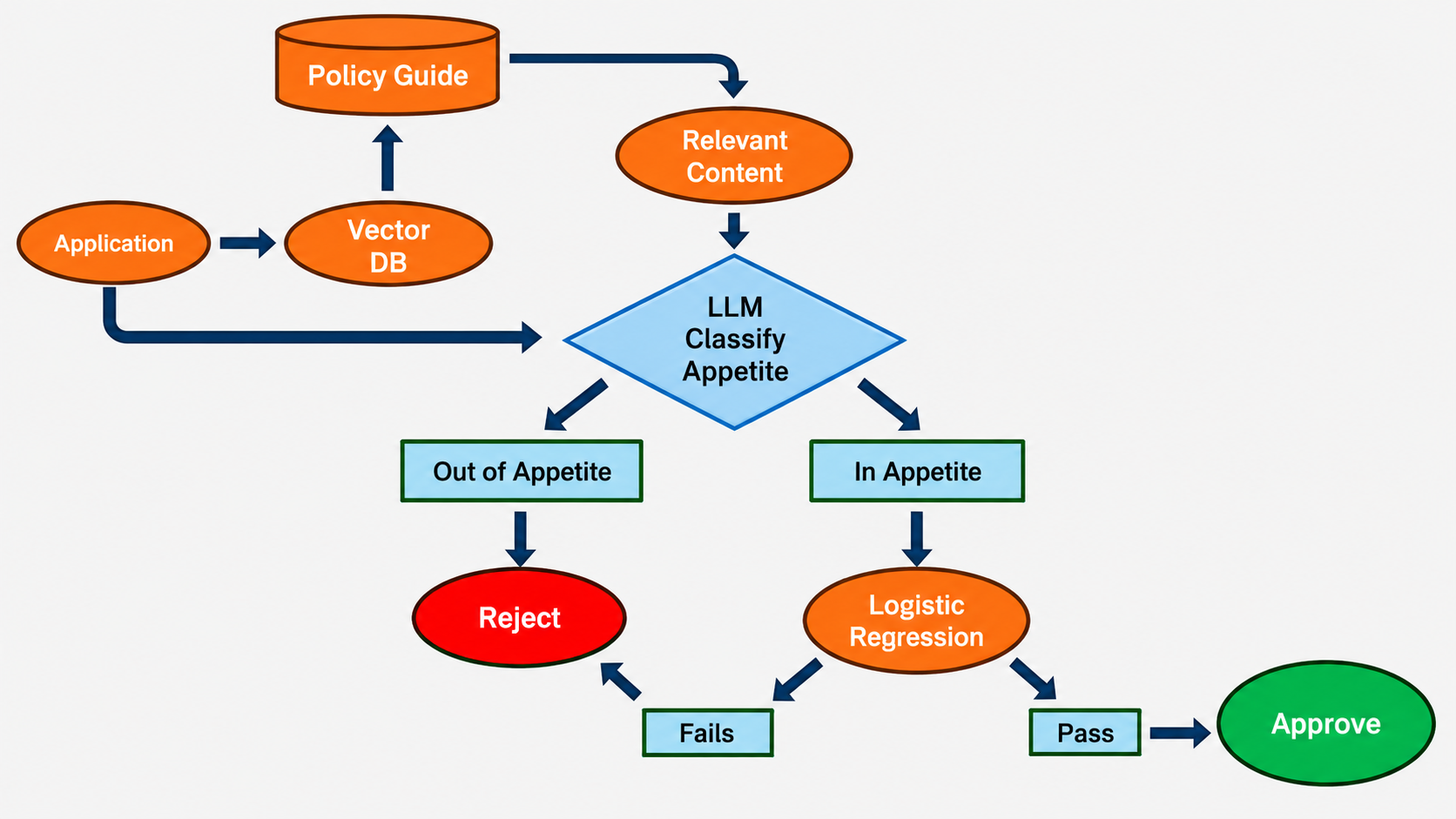}
\caption{\textbf{RAG Workflow.} An LLM receives both the application and retrieved guidebook context. This enables evidence-grounded appetite classification, though the process remains single-pass and lacks explicit diagnostic, completeness, or reflective feedback.}
\label{fig:rag_workflow}
\end{figure}

Figure~\ref{fig:rag_workflow} extends the baseline by adding a retrieval-augmented generation layer that supplements each application with relevant passages from the synthetic underwriting guidebook.

The RAG system improves factual grounding by retrieving domain-specific underwriting rules prior to model inference. However, retrieval alone does not guarantee that the model will identify the decisive missing fact, combine facts correctly, or recognize that a decision should be referred rather than forced.

\subsubsection{Agentic Pipeline}

\begin{figure*}[t]
\centering
\includegraphics[width=\textwidth]{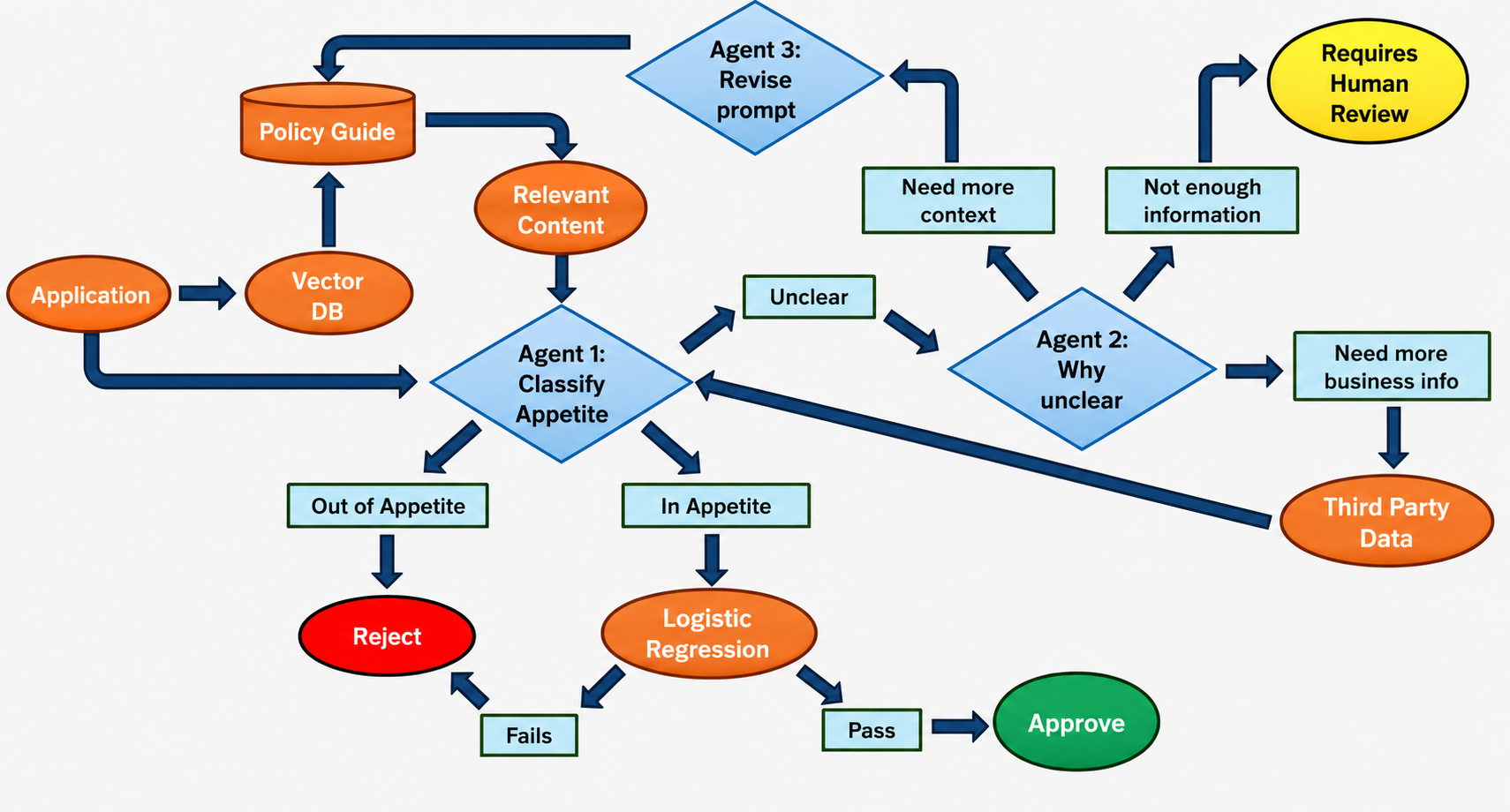}
\caption{\textbf{Agentic Multi-Agent Pipeline.} A coordinated three-agent system with targeted retrieval and reflection loops. Agent~1 performs initial appetite routing, Agent~2 diagnoses missing or ambiguous information and retrieves third-party data when available, and Agent~3 checks multi-step logic before returning the case for reevaluation. Resolved cases are accepted or rejected; unresolved cases are referred for human review.}
\label{fig:agentic_workflow}
\end{figure*}

Figure~\ref{fig:agentic_workflow} illustrates the complete agentic straight-through underwriting pipeline, composed of three cooperating agents orchestrated through a stateful workflow. Each agent specializes in a distinct task: routing, clarification, and reflection.

The sequence proceeds as follows:

\begin{description}
\item[\textbf{Agent~1: Routing and Appetite Check.}] \textcolor{blue}{Evaluates whether the application appears to be \emph{in appetite}, \emph{out of appetite}, or \emph{unclear}. Direct violations are rejected, clear compliant cases are accepted, and uncertain cases are passed to Agent~2.}
\item[\textbf{Agent~2: Contextual Clarification.}] \textcolor{blue}{Determines why uncertainty exists. If the missing element is a guidebook rule, the agent requests targeted retrieval. If the missing element is a business fact, the agent searches the simulated third-party data. If the decisive fact remains unavailable, the case is marked for human review.}
\item[\textbf{Agent~3: Multi-Step Reflection.}] \textcolor{blue}{Checks whether the case requires derived reasoning, such as allocating square footage, combining employee and subcontractor counts, converting revenue percentages into activity-specific revenue, or linking a shared-premises note to an exclusion. The agent synthesizes the relevant facts and returns a structured rationale to Agent~1 for final evaluation.}
\end{description}

This design creates an explicit pathway for deciding when the application contains enough information to support a straight-through decision and when it should be referred. The value of the agentic architecture is not simply that it retrieves more text, but that it organizes the underwriting task into completeness, evidence, and rule-application checks.

\subsection{Implementation Details}
\label{sec:impl}

\textcolor{blue}{All three pipelines were implemented in Python~3.12 using the OpenAI API and the LangChain ecosystem, including \texttt{langchain}, \texttt{langchain-openai}, \texttt{langchain-community}, and \texttt{langgraph}. No local model inference was used. We evaluated two foundation models: \texttt{gpt-4o-mini}, used as a smaller lower-cost model, and \texttt{gpt-5.2}, used as the more capable model. The same pipeline code was used for both models, with only prompt calibration adjusted across model families. Evaluation calls used \texttt{temperature}$=0$ to reduce randomness, while synthetic data generation used a higher temperature to encourage diversity in the generated applications.}

\textcolor{blue}{The underwriting guidebook was parsed into a FAISS vector index for retrieval. We used \texttt{PyPDFLoader} to extract guidebook text and a \texttt{RecursiveCharacterTextSplitter} with 1000-character chunks and 200-character overlap. The na\"{i}ve RAG pipeline retrieved the top guidebook chunks from a single query built from the business type and operations description. The Agentic RAG pipeline allowed additional targeted retrieval when an application was unclear, including queries tied to the business type, global exclusions, and any specific missing or ambiguous issue identified during the intermediate review step.}

\textcolor{blue}{Prompt design was intentionally structured so that routing decisions could be audited. Each model was instructed to return a short JSON object containing a decision label, a rationale, and any missing or decisive underwriting facts. The single-LLM baseline used one prompt containing the application and static guidebook context. The na\"{i}ve RAG model used one prompt containing the application and retrieved guidebook excerpts. The Agentic RAG workflow used a LangGraph state machine with separate steps for appetite review, missing-information diagnosis, targeted retrieval or third-party lookup, multi-step rule evaluation, and final decision review. If a response could not be parsed reliably, the system defaulted to human review rather than forcing an accept or reject decision.}

\textcolor{blue}{To reduce label leakage, each application record was sanitized before evaluation. The model-facing record excluded the scenario label, ground-truth decision, generation rationale, and internal validation fields. The pipelines therefore observed only the application fields, operations narrative, relevant guidebook context, and simulated third-party data. The released artifacts include the synthetic application dataset, per-application evaluation results, pipeline code, and a data dictionary for the application fields. The main implementation notebook is available at} \url{https://github.com/drbob-richardson/agentic-bop-underwriting/blob/main/Agentic_RAG_Underwriting.ipynb}\textcolor{blue}{.}

\subsection{Experimental Setup}

\textcolor{blue}{We tested the framework on the synthetic dataset described in Section~4, covering 127 business types with five scenario categories each: compliant, single-issue violation, multi-step reasoning, missing information recoverable, and missing information irrecoverable. This produces a total of 635 applications.}

\textcolor{blue}{Three configurations were evaluated:}
\begin{enumerate}
\item \textcolor{blue}{\textbf{LLM-Only:} A baseline large language model making single-pass decisions without dynamic retrieval.}
\item \textcolor{blue}{\textbf{Na\"{i}ve RAG:} A standard retrieval-augmented model that retrieves guidebook context but does not use agentic orchestration.}
\item \textcolor{blue}{\textbf{Agentic RAG:} A multi-agent framework with retrieval, completeness checking, third-party data access, and reflection over multi-step logic.}
\end{enumerate}

\textcolor{blue}{Each model was asked to provide both a decision label and a textual justification. These were compared to the human-validated ground-truth outcomes and reasons. Evaluation metrics include:}
\begin{itemize}
\item \textcolor{blue}{\textbf{Decision accuracy:} the percentage of correctly classified outcomes.}
\item \textcolor{blue}{\textbf{Per-scenario accuracy:} accuracy within each of the five scenario categories.}
\item \textcolor{blue}{\textbf{Rationale similarity:} mean cosine similarity between the model-generated explanation and the human-validated ground-truth reason, computed across all 635 applications for each pipeline.}
\item \textcolor{blue}{\textbf{Latency:} average time required to process an application.}
\end{itemize}

\textcolor{blue}{To prevent label leakage, every record is stripped of its scenario tag, ground-truth label, and generation rationale before any pipeline observes it. The pipelines see only the application form, retrieved guidebook context where applicable, and simulated third-party data where the workflow permits it. We report each pipeline under two base models: a small, inexpensive model (\texttt{gpt-4o-mini}) and a substantially more capable model (\texttt{gpt-5.2}).}

\subsection{Experimental Results}

\textcolor{blue}{Table~\ref{tab:overall} reports overall decision accuracy. Under both base models the Agentic RAG pipeline attains the highest accuracy. The gains are especially meaningful because the revised dataset is designed to test not only direct rule recognition, but also multi-step rule application and appropriate referral when decisive information is unavailable.}

\begin{table}[h]
\centering
\caption{Overall decision accuracy (\%) against human-validated ground truth, $n=635$. Best per column in bold.}
\label{tab:overall}
\begin{tabular}{lcc}
Pipeline & \texttt{gpt-4o-mini} & \texttt{gpt-5.2} \\
Single-LLM & 73.4 & 77.6 \\
Na\"{i}ve RAG & 71.5 & 76.9 \\
Agentic RAG & \textbf{85.1} & \textbf{86.5} \\
\end{tabular}
\end{table}

\subsection{Per-Scenario Accuracy}
\label{sec:perscenario}

\textcolor{blue}{Table~\ref{tab:perscenario} decomposes accuracy by scenario for the \texttt{gpt-5.2} configuration. The agentic pipeline performs best on four of the five scenario categories and is only slightly behind the single-LLM baseline on single-issue violations. The largest practical gains occur in the multi-step and missing-information settings, where a single-pass model is more likely to overlook a derived fact or force a decision when the evidence is incomplete.}

\begin{table}[t]
\centering
\caption{Per-scenario decision accuracy (\%) under \texttt{gpt-5.2}. Best per row in bold.}
\label{tab:perscenario}
\begin{tabular}{lccc}
Scenario & Single-LLM & Na\"{i}ve RAG & Agentic RAG \\
Compliant (accept)           & 96.9 & 95.3 & \textbf{100.0} \\
Single-issue violation       & \textbf{84.3} & 83.5 & 81.9 \\
Multi-step Reasoning         & 70.1 & 66.1 & \textbf{78.0} \\
Missing info (recoverable)   & 80.3 & 85.8 & \textbf{88.2} \\
Missing info (irrecoverable) & 56.7 & 53.5 & \textbf{84.3} \\
\end{tabular}
\end{table}

\textcolor{blue}{The multi-step reasoning row is particularly important because it replaces the earlier logistic-failure scenario. These cases are not failures of quantitative scoring; they are failures or successes of document-grounded logic. The model must identify which facts matter, combine them, and then apply the guidebook rule. The agentic pipeline improves this category from 70.1\% for the single-LLM baseline and 66.1\% for na\"{i}ve RAG to 78.0\%.}

\textcolor{blue}{The missing-information rows show a related pattern. When information is recoverable, the agentic pipeline benefits from an explicit process for seeking third-party data. When information is irrecoverable, the agentic pipeline is much more likely to recognize that the application should be referred rather than forcing an unsupported accept or reject.}

\subsection{Rationale Similarity and Latency}
\label{sec:rationale-latency}

\textcolor{blue}{Table~\ref{tab:rationale_latency} reports two secondary diagnostics under \texttt{gpt-5.2}: rationale similarity and latency. Rationale similarity is an overall pipeline-level metric, not a scenario-specific metric and not a metric restricted to applications referred for human review. For each of the 635 applications, the model-generated rationale is compared with the corresponding human-validated ground-truth rationale using embedding-based cosine similarity. The table reports the mean of these 635 similarities for each pipeline. Higher values indicate that the model's explanation is semantically closer to the human-validated reason for the decision.}

\begin{table}[h]
\centering
\caption{Mean rationale--ground-truth cosine similarity and mean latency under \texttt{gpt-5.2}. Rationale similarity is averaged across all 635 applications for each pipeline.}
\label{tab:rationale_latency}
\begin{tabular}{lcc}
Pipeline      & Rationale cos. & Latency (s) \\
Single-LLM    & 0.709 & 1.64 \\
Na\"{i}ve RAG & 0.741 & 2.14 \\
Agentic RAG   & \textbf{0.779} & 4.72 \\
\end{tabular}
\end{table}

\textcolor{blue}{The agentic pipeline has the highest mean rationale similarity, increasing from 0.709 for the single-LLM baseline and 0.741 for na\"{i}ve RAG to 0.779. This suggests that the agentic workflow not only improves decision accuracy but also produces explanations that are more closely aligned with the human-validated rationales. The metric should still be interpreted cautiously: cosine similarity measures semantic proximity, not actuarial correctness, legal sufficiency, or regulatory adequacy. It is therefore best viewed as a supporting diagnostic alongside decision accuracy and per-scenario accuracy.}

\textcolor{blue}{The improvement in rationale similarity comes with higher latency. The single-LLM baseline averages 1.64 seconds per application, na\"{i}ve RAG averages 2.14 seconds, and Agentic RAG averages 4.72 seconds. This reflects the additional retrieval, third-party checks, and multi-step review used by the agentic workflow. In deployment, the depth of retrieval and reflection should be tuned to the cost of delay, the cost of unnecessary referral, and the risk of an incorrect straight-through decision.}

\subsection{Discussion}
\label{sec:discussion}

\color{blue}

\paragraph{The agentic advantage is concentrated in complex and incomplete cases.}
Overall accuracy favors the Agentic RAG pipeline under both base models: 85.1\% under \texttt{gpt-4o-mini} and 86.5\% under \texttt{gpt-5.2}. The per-scenario results explain why. The agentic system is not uniformly better on every simple case; the single-LLM baseline is slightly stronger on single-issue violations. However, the agentic pipeline is strongest on compliant accepts, multi-step reasoning, recoverable missing information, and irrecoverable missing information. This is the pattern one would hope to see from an architecture designed around retrieval, clarification, and reflection.

\paragraph{Retrieval alone is not enough.}
The naive RAG system improves access to guidebook text, but it does not consistently improve decisions. In the multi-step reasoning scenario, naive RAG performs worse than the single-LLM baseline (66.1\% versus 70.1\%). This suggests that the limiting factor is not merely whether the relevant rule is present in context. The system must also know what facts to extract, what intermediate quantity to compute, and when the available evidence is insufficient. Retrieval supplies information; agentic orchestration supplies a workflow for using that information.

\paragraph{Missing information requires an explicit escalation pathway.}
The irrecoverable missing-information category shows the clearest architectural difference. The correct response in these cases is not to guess; it is to refer the application for human review. The Agentic RAG pipeline reaches 84.3\% accuracy in this category, compared with 56.7\% for the single-LLM baseline and 53.5\% for naive RAG. This result supports the central claim of the paper: agentic structure is most valuable when the underwriting task requires recognizing the limits of the available evidence.

\paragraph{The accept/reject/refer boundary remains a policy choice.}
The results should not be interpreted as showing a single universally optimal automation threshold. In practice, the decision to accept, reject, or refer depends on the relative costs of false acceptance, false rejection, unnecessary human review, and customer delay. A conservative insurer may prefer more referrals, while a high-volume digital channel may prefer a narrower referral boundary. The agentic framework is useful because it makes this boundary explicit and adjustable rather than burying it inside a single prompt response.

\paragraph{Rationale quality improves with structure, but latency increases.}
The agentic pipeline produces the highest rationale similarity under \texttt{gpt-5.2}, with a mean cosine similarity of 0.779. This supports the idea that structured intermediate reasoning can improve not only the final decision but also the explanation attached to that decision. The cost is latency: the agentic workflow averages 4.72 seconds, compared with 1.64 seconds for the single-LLM baseline and 2.14 seconds for naive RAG. For straight-through underwriting, this additional cost may be acceptable for difficult cases but unnecessary for obvious accepts or explicit violations. This motivates future work on adaptive reflection, where the system invokes the full agentic loop only when the expected value of additional reasoning exceeds its cost.

\paragraph{Limitations.}
The corpus is synthetic, and the guidebook rules are stylized. Although the scenarios were manually reviewed for consistency and realism, they cannot fully represent the ambiguity, noise, and institutional variation of production underwriting. The multi-step cases are designed to test a specific form of logical composition, but real underwriting may involve more subtle judgment, conflicting evidence, regulatory constraints, or broker-specific context. In addition, the reported results use a fixed retrieval setup and prompt design. Production deployment would require calibration against real underwriter decisions, monitoring for drift, and careful governance around what the system is allowed to decide automatically.
\color{black}

\section{Conclusion and Future Work}

This study demonstrates that agentic artificial intelligence can provide a coherent framework for automating and enhancing selected parts of the underwriting process in small commercial insurance. By integrating retrieval, completeness checking, third-party data access, and reflection over multi-step rules, the system mirrors realistic decision pathways more closely than a single-pass LLM or a naive RAG workflow. The revised experiment removes the earlier logistic-regression-style simulated screen and instead focuses on a more document-grounded problem: determining whether the correct underwriting decision can be reached by connecting facts across the application, third-party information, and the guidebook.

The results suggest that the agentic approach is most valuable in the cases where straight-through underwriting is most fragile. Simple accepts and direct violations can often be handled by simpler systems. The harder cases involve derived quantities, missing facts, or uncertainty about whether the available evidence is sufficient. In those settings, the agentic pipeline improves accuracy and produces more aligned explanations, while also providing a clearer audit trail.

In practice, the full value of an agentic pipeline is realized when AI and humans work together. The system can reduce cognitive load by handling straightforward cases, retrieving relevant context, and documenting intermediate reasoning. Human reviewers benefit from structured explanations and clearer uncertainty signals, while retaining authority over nuanced or high-impact decisions. This is consistent with actuarial standards of professionalism, transparency, and accountability.

\textcolor{blue}{Several limitations should be emphasized. The experiment uses a synthetic BOP dataset rather than production underwriting data. Although the revised dataset includes missing-information cases and multi-step reasoning scenarios, it does not capture the full complexity of real insurer data, policy forms, legacy systems, or operational constraints. The guidebook and third-party data are also simulated, so the results should be interpreted as a controlled comparison of workflow designs rather than a production validation study. In addition, the results depend on the prompts, retrieval setup, and foundation model used in the experiment. These choices would need to be recalibrated in a different underwriting environment. Most importantly, the system is intended to support triage and documentation, not to replace underwriter or actuarial judgment.}

A key direction for future research lies in the decision economics of agentic workflows. Every retrieval or reflection call carries an implicit cost in latency, compute expenditure, and potentially external data licensing fees. Optimally, an agent should weigh the expected value of improved certainty against these marginal costs. Formalizing this trade-off could yield an economic control model of reasoning depth, where the decision to requery, retrieve third-party data, or invoke reflection is governed by an expected utility function.

\textcolor{blue}{Future work should also compare retrieval-based systems with fine-tuned language models. Fine-tuning may be useful for stable extraction, routing, or output-formatting tasks, and may reduce latency or cost when smaller local models are sufficient. However, RAG is better suited to underwriting rules that change over time or require auditable citation to current guidebook language. Our preliminary fine-tuning experiments did not show a clear improvement over the foundation-model baselines, but they were not broad enough to support a formal conclusion. A likely production approach is a hybrid system that uses fine-tuning for stable recurring tasks and RAG for dynamic context and evidence retrieval.}

Future work will also focus on empirical validation in partnership with insurers and insurtech firms. Synthetic data are valuable for controlled experimentation, but true robustness requires evaluation on heterogeneous, production-grade submissions. Collaborative pilots can test the system's reliability under real ingestion conditions such as missing fields, OCR noise, ambiguous narrative text, conflicting third-party signals, or changing underwriting appetite.

Several technical enhancements remain open for exploration:
\begin{itemize}
  \item \textbf{Adaptive reflection loops:} dynamically deciding when to terminate reasoning based on confidence thresholds and cost penalties.
  \item \textbf{Context optimization:} selective retrieval that balances semantic relevance with token economy.
  \item \textbf{Robust data ingestion:} improving resilience to unstructured or noisy input using hybrid symbolic-neural preprocessing.
  \item \textbf{Human-AI collaboration models:} formalizing escalation triggers and accountability boundaries between automated and actuarial review stages.
  \item \textbf{Cross-domain generalization:} extending beyond BOP to personal lines, reinsurance analytics, and claims reserving while maintaining interpretability.
\end{itemize}

Agentic AI offers a path toward structured, economically aware automation. By aligning computational effort with actuarial principles of efficiency, fairness, and auditability, such systems can evolve from experimental prototypes into operational decision partners. The long-term vision is not to replace human underwriters or actuaries, but to strengthen their ability to supervise, calibrate, and continually improve modern insurance decision workflows.

\section*{Acknowledgements}

The author gratefully acknowledges the contributions of the students who provided human oversight during the evaluation of model outputs. Their careful review, structured feedback, and thoughtful discussions substantially improved the realism and reliability of the agentic AI framework developed in this study. In alphabetical order, sincere thanks go to {Breanne Bair}, {Nathan Brueck}, {Vincent Galbraith}, {Cade Sanders}, and {Riley Williams}. Their work strengthened both the empirical rigor and the practical relevance of this research.

\bibliographystyle{apalike}
\bibliography{agentic_ai_references}

\appendix
\color{blue}
\section{Construction of Multi-Step Reasoning Scenarios}
\label{app:multistep}

The revised experiment replaces the earlier logistic-regression-style quantitative screen with a set of multi-step underwriting scenarios. These scenarios are designed to test whether a model can combine facts across documents before applying an underwriting rule.

A typical multi-step case has four components:
\begin{enumerate}
  \item \textbf{A guidebook rule} that defines a threshold, exclusion, or eligibility condition.
  \item \textbf{An application fact} that provides part of the information needed to apply the rule.
  \item \textbf{A second application or third-party fact} that must be connected to the first fact.
  \item \textbf{A derived conclusion} that determines whether the application should be accepted, rejected, or referred.
\end{enumerate}

The scenarios were generated so that the decisive issue is not usually stated in a single sentence. Instead, the model must identify the relevant rule, determine what quantity or condition the rule requires, extract the needed facts, and apply the rule to the derived value. The following templates were used repeatedly across business types.

\paragraph{Square-footage allocation.}
The application states total leased square footage, while a separate note states the percentage devoted to a specific operation. The guidebook threshold applies only to that operation. For example, if a business leases 14,000 square feet and 50\% is used for a grocery operation, the derived grocery area is 7,000 square feet. If the guidebook permits grocery operations only below 5,000 square feet, the correct decision is reject.

\paragraph{Revenue-mix thresholds.}
The application states total revenue and the percentage generated by a restricted activity. The guidebook may exclude a risk if more than a certain share of revenue comes from that activity. The model must connect the revenue share to the relevant exclusion and determine whether the threshold is exceeded.

\paragraph{Shared-premises conditions.}
The application may describe an eligible business, while a lease note or third-party description indicates that the business shares premises with an excluded operation. The model must recognize that the shared-premises rule applies even when the primary operation is otherwise eligible.

\paragraph{Combined worker counts.}
Some rules apply to total on-site workers, while the application lists employees, contractors, and seasonal workers separately. The model must combine the relevant counts before applying the guidebook threshold.

\paragraph{Activity-specific operations.}
A business may be eligible for its main operation but ineligible if a secondary activity exceeds a stated frequency, area, or revenue threshold. For example, a studio may be acceptable for standard instruction but require referral if it hosts high-attendance events more than a specified number of times per month.

These examples are intentionally simple enough to permit human validation, but complex enough to distinguish rule lookup from structured rule application. They also create a natural role for agentic orchestration: one agent can retrieve the rule, another can identify missing or derived facts, and a reflection step can check whether the final decision follows from the evidence.
\color{black}

\end{document}